\documentclass[10pt,twocolumn,letterpaper]{article}
\usepackage{iccv}
\usepackage{times}
\usepackage{epsfig}
\usepackage{wrapfig}
\usepackage{soul}
\usepackage{amsmath}
\usepackage{amssymb}
\usepackage{microtype}
\usepackage{wrapfig}
\usepackage{tabularx}
\usepackage{graphicx}
\usepackage{booktabs}
\usepackage{xcolor}
\usepackage[utf8]{inputenc}
\usepackage{multirow, makecell}

\def\figurePath{images/}
\def\myfigure#1#2{\begin{figure}[h]\centering\includegraphics*[width = \linewidth]{\figurePath#1}\caption{#2}\label{fig:#1}\end{figure}}

\def\myfiguretop#1#2{\begin{figure}[t]\centering\includegraphics*[width = \linewidth]{\figurePath#1}\caption{#2}\label{fig:#1}\end{figure}}

\def\mycfigure#1#2{\begin{figure*}[tb]\centering\includegraphics*[clip, width = \linewidth]{\figurePath#1}\caption{#2}\label{fig:#1}\end{figure*}}

\renewcommand{\eg}{e.\,g., }
\renewcommand{\ie}{i.\,e., }
\renewcommand{\etal}{et~al.\ }

\newcommand{\argmin}[1]{\underset{#1}{\operatorname{arg\,min}\ }}

\newcommand{\refSec}[1]{Sec.~\ref{sec:#1}}
\newcommand{\refFig}[1]{Fig.~\ref{fig:#1}}

\newcommand{\refTbl}[1]{Tbl.~\ref{tbl:#1}}

\soulregister\ref7
\soulregister\cite7
\soulregister\refFig7
\soulregister\cite7
\soulregister\ref7
\soulregister\pageref7
\soulregister\shortcite7
\soulregister\eg0
\soulregister\ie0
\soulregister\etal0

\DeclareGraphicsExtensions{.png,.jpg,.pdf,.ai,.psd}
\newcommand{\mysection}[2]{\section{#1}\label{sec:#2}}
\newcommand{\mysubsection}[2]{\subsection{#1}\label{sec:#2}}

\iccvfinalcopy

\newcolumntype{s}{>{\hsize=.8\hsize}X}
\newcommand{\heading}[1]{\multicolumn{1}{c}{#1}}

\definecolor{darkgreen}{rgb}{0,.4,0}

\usepackage[pagebackref=true,breaklinks=true,letterpaper=true,colorlinks,bookmarks=false]{hyperref}

\iccvfinalcopy 

\def\iccvPaperID{4021}

\ificcvfinal\fi
\begin{document}

\title{Deep Appearance Maps}

\author{
Maxim Maximov
\vspace{.2cm}\\
Technical University\\Munich
\and
Laura Leal-Taixé
\vspace{.2cm}\\
Technical University\\Munich
\and
Mario Fritz
\vspace{.2cm}\\
CISPA Helmholtz Center\\for Information Security
\and
Tobias Ritschel
\vspace{.2cm}\\
University College\\London
}
\maketitle
\ificcvfinal\thispagestyle{empty}\fi
\begin{abstract}
We propose a deep representation of appearance, \ie the relation of color, surface orientation, viewer position, material and illumination.
Previous approaches have used deep learning to extract classic appearance representations relating to reflectance model parameters (\eg Phong) or illumination (\eg HDR environment maps).
We suggest to directly represent appearance itself as a network we call a Deep Appearance Map (DAM).
This is a 4D generalization over 2D reflectance maps, which held the view direction fixed.
First, we show how a DAM can be learned from images or video frames and later be used to synthesize appearance, given new surface orientations and viewer positions.
Second, we demonstrate how another network can be used to map from an image or video frames to a DAM network to reproduce this appearance, without using a lengthy optimization such as stochastic gradient descent (learning-to-learn).
Finally, we show the example of an appearance estimation-and-segmentation task, mapping from an image showing multiple materials to multiple deep appearance maps.
\end{abstract}

\mysection{Introduction}{Introduction}
The visual appearance of an object depends on the combination of four main factors: viewer, geometry, material and illumination.
When capturing and processing appearance, one wishes to change one or more of those factors and predict what the new appearance is.
This can be achieved using methods ranging from implicit image-based representations \cite{debevec1996modeling,horn1979calculating} to explicit Computer Graphics-like representations \cite{pharr2016physically}.
Implicit methods take a couple of photos as input and allow to predict high-quality imagery in a limited set of conditions, but modest flexibility, \eg interpolating an image between two photos but not extrapolating to new views.
Explicit representations allow for more flexibility when acquiring Phong parameters and HDR illumination maps \cite{pharr2016physically}, but incur substantial acquisition effort, \eg taking a large number of calibrated (HDR) photos.

DAMs propose a new direction to represent appearance: we move away from the pixel-based nature of implicit image-based representations into a deep representation, but without any explicit reconstruction, as we do not target a direct mapping to any explicit reflectance model or illumination either.
Still, we show that such a representation can be used to solve actual tasks, such as image synthesis, appearance acquisition and estimation-and-segmentation of appearance.
This is enabled by four contributions:

First, we will introduce a generalization of 2D reflectance maps \cite{horn1979calculating} to 4D, which we call an Appearance Map (AM).
AMs represent appearance for varying geometry under varying views.
This allows freely changing the viewer and surface geometry, which is not possible for classic reflectance maps that fix the relation between view and illumination (cf. \refFig{Teaser}).

\myfiguretop{Teaser}{
Frames from a video with a moving viewer \textbf{(columns)} comparing
a re-synthesis using our novel deep appearance maps (DAMs) \textbf{(top)} and reflectance maps (RMs) \textbf{(bottom)} to a photo reference of a decorative sphere with a complex material under natural illumination \textbf{(middle)}.
}

Second, while classic Reflectance Maps (RM) can be simply tabulated using a single 2D image, the full appearance is a 4D phenomenon that is more challenging to represent and process.
Storing 4D appearance as a table, modestly resolving 10 degree features, would require storing (and somehow also capturing) $36^4=17\,M$; impractical.
Instead, we suggest using DAMs, neural networks that compactly represent AMs in only a couple of thousand parameters.
This representation is efficient, does not require any look-ups and is differentiable. 
In addition, it can be learned effectively from images or video frames with a known viewer position and surface orientation.
Applying this representation to new view positions and surface orientations can be done at speed competitive to classic rendering or RMs, \ie within milliseconds for full images.

Acquiring a DAM requires learning a deep model for every new appearance in practice.
This would incur substantial computational effort, \ie running an optimization compared to capturing a RM image in seconds.
Addressing this, our third contribution suggests to use another deep (convolutional) neural network to map images showing an appearance to a DAM (\cite{lake2017building,thrun2012learning,jia2016dynamic}, one-shot learning \cite{fei2006one}, ``life-long'', or ``continual'' learning).
This capture requires milliseconds instead of minutes.

Fourth, the DAM representation can be used for joint material estimation-and-segmentation, a generalization of the previous objective.
Here the input is an image with a known number of $n$ materials, and output is $n$ different DAMs, and a segmentation network that maps every pixel to a $n$ weights.

We train and quantitatively test all networks on a new dataset of photo-realistically rendered images as well as on a set of real photos.

\mysection{Related Work}{RelatedWork}

\paragraph{Inverse Rendering}
One of the main aims of inverse rendering is to recover material and illumination properties of a scene.
It is a quite challenging, ill-posed and under-constrained problem that remains hard to solve for the general case.  
Related recent work can be roughly divided into data-driven and algorithmic approaches.

Algorithmic methods are based on optimizing appearance properties for a given input \cite{marschner1999image}.
These methods are usually off-line and make simplifying assumptions about the world to reduce computation time and avoid ambiguity and allow for a mathematical derivation.
Most recent works \cite{Wu2016,RichterTrummer2016} use a set of real RGBD images to estimate appearance that are based on a specific illumination model. 
More refined models use data-based statistical priors to optimize for illumination and reflectance explaining an image \cite{matusik2003data,lombardi2012reflectance}.

Deep-learning based approaches make a similar assumption as to how humans can recognize materials based on previous experience.
Recent work \cite{MekaA2018,Georgoulis2017Around,Liu2017,Kim2017,deschaintre2018singleimage} uses CNNs to estimate explicit reflectance model parameters.
Similarly, encoder-decoder CNNs are used to estimate reflectance maps \cite{Rematas2016} or illumination and materials \cite{hold2017deep,Georgoulis2017Around}.

All of theses methods -- data-driven or not -- have in common that they rely on a specific illumination model to estimate its explicit parameters (such as Phong diffuse, specular, roughness, etc) and they represent lighting as an HDR illumination map.
To the one hand, this is more that what we do as it factors out lighting, to the other hand our approach is more general as it makes no assumption on light or geometry and works on raw 4D samples.
One of the other limitations of above mentioned CNN methods is limited feedback from a loss function: a change of estimated illumination or reflectance can only be back-propagated through the image synthesis method with suitable rendering layers.
Our method does not involve a renderer, circumventing this problem.

\paragraph{Appearance synthesis}
Methods to synthesize appearance -- or simply ``rendering'' --, can be classified as simulation-based or image-based.

Simulation-based methods require a complete explicit description of the environment that can be costly and difficult to acquire in practice \cite{pharr2016physically}.
A simple, yet powerful, method to represent appearance is a reflectance map \cite{horn1979calculating}, a 2D image that holds the color observed for a surface of a certain orientation and material under a certain illumination.
In graphics, reflectance maps are known as pre-filtered environment maps \cite{kautz2000unified}, or spherical harmonics (SH) to capture the entire light transport \cite{sloan2002precomputed}.
A single 2D envmap or 2D SH however, cannot reproduce 4D appearance variation and furthermore requires a high pixel resolution or many SH coefficients to capture fine details (highlights) that are easy to reproduce using a NN.
The entire reflectance field is a spatial map of these, as captured by Debevec \cite{debevec2000acquiring}.

Image-based rendering  (IBR) uses a set of 2D images to reconstruct a 3D model and present it in a different view or different light \cite{debevec1996modeling}.
These methods do geometry prediction, often with manual intervention, with prediction of rendered material on top of it. 
Recent methods \cite{zhou2016view,Rematas2017} address this problem by using CNN models to predict completely novel views.
The method of Rematas~\etal\cite{Rematas2016} and establish a relation between surface and light transport properties and appearance given by photos, generating images ``without rendering''.
A simple data-driven approach to IBR is to learn a per-pixel fully-connected neural network to reproduce per-pixel appearance \cite{ren2015image} depending on light.
A generalization of this is to shade images with per-pixel positions, normals and reflectance constraints \cite{nalbach2017deepshading}.
Our method stems from the same root but neither works on pixel-based image rendering, nor does it reconstruct an explicit appearance model.
We will instead use a deep representation of appearance itself.

Light fields \cite{levoy1996light} also store 4D appearance, but are parametrized by spatial or surface position \cite{wood2000surface} and store 2D lumitexels.
BRDFs 4D-capture reflectance, but not illumination.
Our DAMs capture the 4D relation of surface orientation and view direction instead.

\paragraph{Segmentation}
Classic segmentation does not take materials into account \cite{shi2000normalized}.
Recent material segmentation work, such as Bell~\etal\cite{bell2015material} is mostly a form of extended semantic segmentation into material labels (23 in their case): Most arm chairs might be made of only three different kinds of materials that such approaches are successful in detecting.
In our work, we have abstract objects (like photographs of spheres \refFig{Teaser}), that do not provide much semantics and require using a continuous appearance representation.
For videos of view-dependent appearance, this is particularly difficult.
With adequate capture equipment, spatially varying appearance is captured routinely now \cite{lensch2003image,goldman2010shape}.
In particular, with very dense light field that covers a tiny angular region, changes in appearance can be used to separate specular and diffuse \cite{alperovich2016variational}.
Our work uses much sparser samples and goes beyond a specular-diffuse separation to support arbitrary 4D appearance extrapolated over all view directions.

Another challenge is multi-materials estimation. 
Some work \cite{Georgoulis2017Around,wang2017jointwild} has used multiple materials under the same illumination, but they require pre-segmented materials.
In our method we perform joint multi-material segmentation and estimation.

\paragraph{Learning-to-learn}
Learning-to-learn is motivated by the observation that a general optimizer, such as the one used to find the internal parameters for a network, will never be much better than a random strategy for all problems \cite{wolpert1995no}.
At the same time, intelligent actors can learn very quickly, which obviously does not require a full optimization \cite{lake2017building}.
We hypothesize, after seeing a material for some time, that a human, in particular a trained artist, would be able to predict its appearance in a new condition.
This requires the ability to refine the learned model with new observations \cite{thrun2012learning}.
For convolutional networks, this was done in dynamic filter networks \cite{jia2016dynamic}, but we are not aware of applications to appearance modeling, such as we will pursue here.

\mysection{Deep Appearance Processing}{OurApproach}

\subsection{Appearance maps}
We model RGB appearance $L_\mathrm o$ of a specific material $f_\mathrm r$ under a specific distant illumination $L_\mathrm i$ as a mapping from absolute world-space surface orientation $\mathbf n$ and viewer direction $\omega_\mathrm o$ (jointly denoted as $\mathbf x$) as in \[
L_\mathrm o(
\underbrace{\omega_\mathrm o, \mathbf n}_{=\mathbf x}
)=
\int
L_\mathrm i(\omega_\mathrm i)
f_\mathrm r(\omega_\mathrm i, \omega_\mathrm o)
<\mathbf n, \omega_\mathrm i>^+
\mathrm d \omega_\mathrm i
.
\]
Essentially, $L_\mathrm o$ is a six-dimensional function.
In the following, we denote the two three-dimensional parameters -- outgoing direction $\omega_\mathrm o$ and surface orientation $\mathbf n$ -- as a joint parameter vector $\mathbf x$.
\myfigure{GeneralizedReflectanceMaps}{Reflectance and Appearance maps.}
The concept if visualized in \refFig{GeneralizedReflectanceMaps}: In a classic reflectance map, the normals vary (blue arrows), but the view direction is the same (orange arrows).
In our generalization, both view and normals vary arbitrarily.
We might even observe the same normal under different views.
Classic reflectance maps \cite{horn1979calculating}, assume a view direction $\mathbf z$ along the $z$ axis and hold the relation of light and surface fixed, while also being limited to a single half-space:
\[
L_\text{RM}(\mathbf n) \; \text{where} \; < \mathbf n , \mathbf z > \; \le \; \frac{\pi}{2}.
\]
Covering the 4D sphere is motivated by our applications that allows to independently change view (2D) and surface orientation (2D).
Note, no assumption on a BRDF is made as others do \cite{hold2017deep,Georgoulis2017Around}.

\mycfigure{Variants}{
Different appearance processing tasks that we address using our deep appearance maps.
\textbf{a)} The first task simply reproduces a given appearance, \ie it maps from normal and view directions to RGB values using a NN. 
\textbf{b)} In a learning-to-learn task a network maps an image to a DAM representation.
\textbf{c)} Finally, in the segmentation-and-estimation task, a network maps an image to multiple DAMs and multiple segmentation networks.
}

\mysubsection{Deep Appearance Maps}{DeepRepresentation}
We use a deep neural network $\hat L_\mathrm o(\mathbf x|\theta)$ to approximate $L_\mathrm o(\mathbf x)$ where $\theta$ denotes the networks internal parameters (\refFig{Architectures}, a).
The input to such a network is the surface orientation and viewer direction parametrized as Euclidean vectors, \ie a total of six numbers.
This is followed by several fully-connected layers that are ultimately combined into a single RGB output value.
Using $1\times 1$ convolutions provide independence of image or object structure.
Here, stochastic gradient descent (SGD) is used to minimize 
\begin{gather*}
\argmin{\theta,\delta}
c_\mathrm d(\theta,W)
+
\alpha
c_\mathrm a(\theta,\delta)
\end{gather*}
according to the $\alpha$-weighted sum of a data cost $c_\mathrm d$ that depends on the DAM model parameters and an adversarial cost $c_\mathrm a$ that further includes the cost of the parameters of an adversarial model $\delta$, that is biasing the solution to be plausible.
$W$ is a weight vector that is set to 1 for now, but will be required later for segmentation.
We use $\alpha=.001$.
The data cost is defined as
\begin{gather}
c_\mathrm d(\theta)
=
\frac 1 n
\sum_{i=1}^n
W_i
||
\hat L_\mathrm o(\mathbf x_i|\theta)-
L_\mathrm o(\mathbf x_i)
||,
\end{gather}
where 
$L_\mathrm o(\mathbf x)$ are the observed appearance for normal and view direction $\mathbf x$ and
$\hat L_\mathrm o(\mathbf x|\theta)$ is modeled appearance with parameter $\theta$.
The adversarial cost is defined as
\begin{gather}
c_\mathrm a(\theta,\delta)
=
\underbrace{
\sum_{I\in\mathcal I"}
\Delta_\mathrm a(
R(\theta, I'_\mathrm{n/v})
|\delta
)
}_\text{Rendered appearance is fake}
+
\underbrace{
\sum_{I'\in\mathcal I}
(
1-
\Delta_\mathrm a(I'_\mathrm{rgb})
|\delta
)
)
}_\text{Real appearance is real}
,
\end{gather}
where $\mathcal I$ is a set of images with per-pixel color (I\textsubscript{rgb}), normals(I\textsubscript{n}) and view directions (I\textsubscript{b}) (detailed in \refSec{DataSet}),
$\Delta_\mathrm a$ is an adversarial network with parameters $\delta$, classifying its argument as fake when it is 1,
and $R$ is an operator that applies the appearance model with parameters $\theta$ to the per-pixel normals and view directions in image $I$ (re-synthesis / rendering).
The adversarial network $\Delta_\mathrm a$ itself is a common encoder-style 
classifier as detailed in \refFig{Architectures}, b.
It classifies the input image into a single value between 0 and 1.
The GAN incentivizes the solution to produce plausible results in regions where no normals or view directions are observed during training. 

\myfigure{Architectures}{The four architectures used.}

Learning a deep appearance model takes more than a minute while it is executed on a full image within one millisecond.
We will now see how this representation enables two novel applications: learning-to-learn material appearance (\refSec{LearningToLearn}) and material estimation-and-segmentation (\refSec{Segmentation}).

\mysubsection{Learning-to-learn Deep Appearance Models}{LearningToLearn}
Taking it a step further, we suggest to replace the learning procedure explained before by a network (learning-to-learn \cite{thrun2012learning}).
The main idea is to speed up the learning process, allowing acquisition of a deep appearance material on-the-fly at interactive rates (one millisecond) instead of an optimization requiring 71 seconds (for \refTbl{Main} on a Nvidia GTX 1080).
A general optimization process can find a good solution to all problems, but no learning approach that does well on all problems is much better than guessing \cite{wolpert1995no}. 

However, we do not ask for ``free lunch'' here as we know that we do not want to solve all problems, but a specific subset: learning how to map normals and view directions to appearance.

To this end, we employ a convolutional neural network $\Theta(I|\phi)$ that is executed on an image $I$ with a second set of internal parameters $\phi$.
Consequently, a network replaces a general learning algorithm \cite{jia2016dynamic}.
This network can be compactly deployed and efficiently executed on arbitrary images to produce a DAM that can then be used for synthesis. 
In this sense, it is a network that predicts the weights (learned parameters) of network.

The input to the network is a 256$\times$256 RGB, normal and view direction images $I$ showing the appearance of a single material \refFig{Architectures}, c.
Using an image instead of a plain list of samples allows the network to reason about spatial arrangement of values, \eg detecting shapes and relations of highlights.

The output is a 3360-dimensional vector $\Theta(I|\phi)$ that describes the internal parameters of a network producing the appearance of $I$.
The network $\Theta$ has eight layers, reducing resolution until a fully-connected layer.
Training now minimizes for
\begin{gather}
\argmin{\phi, \delta}
\sum_{I\in\mathcal I}
c_\mathrm d(\Theta(I|\phi),\mathbf w)
+
\alpha
c_\mathrm a(\Theta(I|\phi)|\delta),
\end{gather} 
\ie the same cost as in the previous section, but defined on the parameters $\phi$ of a network $\Theta(I|\phi)$ producing another network instead of the network parameters $\theta$.

\mysubsection{Appearance Estimation-and-Segmentation}{Segmentation}
A second application is joint appearance estimation and segmentation.
Instead of holding a segmentation fix and estimating an appearance model for each segment or assuming an appearance to apply a segmentation, we jointly do both in an unsupervised way. 
We suggest using SGD itself as an optimization method to find the segmentation and appearance-predicting networks.
Here, the DAM as well as a segmentation network are used as the latent variables to be inferred.
The number of materials $n$ is assumed to be known. 

The appearance network parameters for all appearances are stacked into a matrix
$
\mathsf\Theta(I)=
(
\Theta(I|\phi_1),
\ldots,
\Theta(I|\phi_n)
)^\mathsf T
$.

Instead of directly inferring a per-pixel segmentation mask in the optimization, we suggest to learn a network $\Psi(\psi_i)$ with parameters $\psi_i$ that jointly produces the all $n$ segmentation masks (\refFig{Architectures}, d). 

This network again is a simple encoder-decoder with skip connections that is shared among the materials in order to further reduce parameters.
Input to this network is an image $I$ with pixel color, normal, position, and output is a weight mask expressing how much a pixel belongs to a certain material $i$.
There is one segmentation network parameter for each material $i$, and they are all stacked into a matrix
$
\mathsf\Psi(I)=
(
\Psi(I|\psi_1),
\ldots,
\Psi(I|\psi_n)
)^\mathsf T
$.
The optimization now becomes
\begin{align}
\argmin{
 \mathsf \Theta,
 \mathsf \Psi,
 \delta_\mathrm a,
 \delta_\mathrm m}
\sum_{i=1}^n
&c_\mathrm d(
 \Theta(I|\phi_i),
 \Psi(I|\psi_i)
)
+\nonumber
\\
\alpha
&c_\mathrm a(
 \Theta(I|\phi_i),
 \delta_\mathrm a)
+
\beta
c_\mathrm s(W) 
.
\end{align}
Here, $c_\mathrm s$ is a sparsity term on the weight mask $W$ that encourages a solution where most values for one pixel are zero, except one \ie to have one unique material in most pixel.
For every channel $\mathbf w$ in $W$ it is $\sum_i \text{abs}(\mathbf w_i-.5)$.

\mysection{A Multi-view Multi-material Dataset}{DataSet}
To work on arbitrary combinations for view, surface orientation and illumination for objects with multiple materials, we first produce a dataset.
To our knowledge, no multi-material, multi-view dataset exists that allows for a controlled study.
Examples from the dataset are shown in \refFig{Dataset}.
\myfigure{Dataset}{
Two samples from four variants of our data set.}

A seemingly obvious way to capture such a dataset from the real world is to take many photos at many exposure settings of many geometric objects under varying illuminations.
Regrettably, this does not scale well to a high number of samples due to the curse of dimensionality encountered with so many parameters (the product of geometry, material, illumination and view).

Also it would be difficult to manually decorate them with ground-truth material segmentation.
Instead, we suggest to use photo-realistically rendered imagery.

We acquired five production-quality 3D objects from a model repository.
As most of our architectures consider images simply as a list of 4D samples, without spatial layout, this comparatively low number of models appears adequate.
Each model was assigned multiple (three) or one physically-plausible (layered GGX shading) materials organized on the objects surface in a complex and natural spatial arrangement.
Before rendering, we randomize the hue of the diffuse component.
For illumination, we used 20 different HDR environment maps.
For each model, 32 different but equidistant view points on a circle around the object, with a random elevation, were used.
Overall, this results in 5$\times$20$\times$32=3200 images.
Note,  the number of photos that would be required to exhaustively cover 4D; an order of magnitude higher.
As the views and materials are randomized, no sharing between test and train sets exists.
Geometry \ie certain combinations of normals and view directions, might occur both in test and training data.
We perform the same split into test and training for all tasks.

For rendering, we use Blender's \cite{blender2018manual} Cycles renderer with high-quality settings, including multiple indirect and specular bounces.
Note that those light paths violate the model assumption.
We add a virtual tripod to be closer to real photos, which typically also have local reflections which invalidate the model assumptions of distant illumination.
The resulting images are linearly tone-mapped such that the 95th~percentile maps to 1 and kept linear (non-gamma corrected).
For each image $I$ in the corpus $\mathcal I$, we store many channels: 
appearance as RGB colors $I_\mathrm c$,
position $I_\mathrm p$,
normals $I_\mathrm n$,
and a weight map $I_\mathrm w$ with $n$ channels, where $n$ is the number of materials.

Additionally to the \textsc{Envmap} version, we produce 
a variant with \textsc{Pointlight} illumination (technically, a single, small area light) and split the set into flavors: \textsc{MultiMaterial} and \textsc{SingleMaterial}.
Using a single material, the material segmentation is ignored and one random material from the objects is assigned to the entire 3D objects.
In the multi-material case, we proceed directly.
Note, that such instrumentation would not be feasible on real photographs.

\mycfigure{Plots}{
Pairs of error plots for each task.
In each pair, the first is the old and the second the new view.
Each curve is produced by sorting the DSSIM (less is better) of all samples in the data set.
Blue colors are for point light illumination, red colors for environment maps.
Dark hues are the competitor and light hues ours.
}

\section{Results}

\mysubsection{Protocol}{Protocol}

Here we evaluate our deep appearance representation (\refSec{RepresentationAnalysis}), as well as its application to learning-to-learn appearance (\refSec{LearningToLearnAnalysis}) and joint material estimation-and-segmentation (\refSec{SegmentationAnalysis}).

Instrumentation for all tasks is performed in a similar fashion using our multi-view, multi-material data set (\refSec{DataSet}).
In particular, we consider its  \textsc{PointLight} and \textsc{ EnvironmentMap} variants.
Depending on the task, we either use a \textsc{SingleMaterial} or \textsc{MultiMaterial}.
The main quantity we evaluate is image similarity error (DSSIM, less is better) with respect to a path-traced reference.
We consider two tasks: re-synthesizing from the \textsc{SameView} (training views) as well as from a \textsc{NovelView} (test views).
We will use 10 of the 32 views for every sample for training and predict 22 novel views.
The 10 views form a random but consecutive range of angles ca.\ 240 degree.

In each application we also consider one application-specific competitor to solve the task.
We use perfect classic reflectance maps for appearance representation \cite{horn1979calculating}, an upper bound with what could be estimated \cite{Rematas2017}.
SGD is the common solution to learn appearance.
For testing on real data, we need to select a single 2D input image for the RM.
We use an oracle that selects the 2D image resulting in the lowest error.
This is an oracle, impossible in practice, as the selection would required knowing the reference.
For material segmentation, no clear baseline exists.
We experimented with the method of Bell~\etal\cite{bell2015material}, but concluded it is trained on semantic shapes (chairs imply wood etc.) which do not transfer to the abstract shapes we study (see supplemental materials for examples).
Therefore, inspired by intrinsic images that also find consistent patches of reflectance \cite{garces2012intrinsic}, we simply employ $k$-means clustering in RGB-Normal space to do joint material estimation-and-segmentation.
We will now look into the three specific applications.

\mysubsection{Appearance representation}{RepresentationAnalysis}
We study how well our approach can represent appearance per-se.
Most distinctly, we propose to use a 4D appearance map while other works use 2D image representations of a reflectance map.
To quantify the difference, we represent the \textsc{SingleMaterial} variant of our dataset as a common reflectance map, as well as using our appearance map.

To emulate a common reflectance map, which is defined in view space, we take the input image from the closest view from the training set as a source image.
Every normal in the new view is converted to camera space of the new view and the same is done for the normal in the old view.
We then copy the RGB value from the old view image to the new-view image that had the most similar normal.
Note, that such a multi-view extension of RMs already is more than the state of the art that would use a single view. 
We call this method RM++.

\refTbl{Main}, top part, shows results as mean error across the data set.
We see that for all data sets our method is a better representation in terms of having a lower DSSIM error.
The difference in error is more pronounced for \textsc{NovelView} than for \textsc{SameView}.
A detailed plot of error distribution is seen in \refFig{Plots}, left.
This is, as classic reflectance map captures appearance for a fixed viewer location, for changing geometry, but does not generalize when the viewer moves.
Arguably, classic RMs look qualitatively plausible without a reference, but only have low quantitative similarity (SSIM) in novel views. 

\begin{table}
\center
\caption{
Quantitative results on synthetic data.
Rows are different combination of tasks and methods (three applications, two view protocols, our two methods).
Columns are different data.
Error is measured as mean DSSIM across the data set (less is better).
}
\label{tbl:Main}
\setlength{\tabcolsep}{3.0pt}
\begin{tabular}{lllrrrrr}
\multirowcell{2}{\heading{Task}}&
\multirowcell{2}{\heading{View}}&
\multirowcell{2}{\heading{Method}}&
\multicolumn{2}{c}{Error}
\\
\cmidrule(lr){4-5}
&&&
\heading{\textsc{Pnt}}&
\heading{\textsc{Env}}
\\
\toprule
\multirowcell{4}{Representation\\(\refSec{DeepRepresentation})}&
\multirowcell{2}{Same}&
\textsc{Our}&
\textbf{.105}&
\textbf{.123}
\\
&
&
\textsc{RM++}&
.143&
.160
\\
&
\multirowcell{2}{Novel}&
\textsc{Our}&
\textbf{.144}&
\textbf{.164}
\\
&
&
\textsc{RM++}&
.181&
.193
\\
\midrule
\multirowcell{4}{Learn-to-learn\\(\refSec{LearningToLearn})}&
\multirowcell{2}{Same}&
\textsc{Our}&
.106&
.131
\\
&
&
\textsc{SGD}&
.105&
.123
\\
&
\multirowcell{2}{Novel}&
\textsc{Our}&
.165&
.173
\\
&
&
\textsc{SGD}&
.144&
.164
\\
\midrule
\multirowcell{4}{Segmentation\\(\refSec{Segmentation})}&
\multirowcell{2}{Same}&
\textsc{Our}&
\textbf{.113}&
\textbf{.122}
\\
&
&
\textsc{kMeans}&
.132&
.136
\\
&
\multirowcell{2}{Novel}&
\textsc{Our}&
\textbf{.161}&
\textbf{.154}
\\
&
&
\textsc{kMeans}&
.172&
.164
\\
\bottomrule
\end{tabular}
\end{table}

\mysubsection{Learning-to-learn appearance models}{LearningToLearnAnalysis}
Here, we also follow the protocol described in \refSec{Protocol}.
After having established the superiority of deep appearance maps to classic reflectance maps in the previous section, we use it as a competitor (SGD) for learning-to-learn.
At best, our learning-to-learn network produces a network which is as good as running a full SGD pass.

\myfigure{LearningToLearnResults}{Results of our DAM representation trained using stochastic gradient descent \textbf{(1st column)}, our DAMs produced by our learning-to-learn network \textbf{(2nd column)} as well as a reference \textbf{(3rd column)} in a novel-view task.}

The middle part of \refTbl{Main} summarizes the outcome when executing the resulting $\phi$ on the test data set.
We see that both approaches reproduce the appearance faithfully.
For point lights, the mean DSSIM is .144 for SGD while it is .165 for network-based (\refTbl{Main}, middle part).
Naturally, letting a network do the learning degrades quality, but only by a marginal amount, in this case 14.1\,\%.
For environment map illumination, the mean DSSIM is increased from .164 to .173, a decrease by only 5\,\%.
While being marginally worse, it is two orders of magnitude faster.
\refFig{Plots} show the error distribution across the data set.

A visual comparison is found in \refFig{LearningToLearnResults}.
We see that replacing the SGD computation of several minutes by a network, can produce a DAM that is qualitatively similar to both the SGD's result as well as to the reference.
Overall, strength and sharpness of highlights that is already challenging for DAM per-se, seems to suffer a bit more by learning-to-learn, as also seen in \refFig{FailureModes}.

\mysubsection{Joint Material Estimation-and-Segmentation}{SegmentationAnalysis}
Finally, we quantify the joint material-and-segmentation task from \refSec{Segmentation}.
We perform the same split as in the previous section, however, now on the \textsc{MultiMaterial} variant.
For the re-synthesis to new views we use the ground truth segmentation in the new view (our method only produces the segmentation in all old views).

We here compare to a competitor, where the image is first segmented using $k$-means clustering on normals and RGB (same weight, as both have a similar range) and material is estimated for each segment in consecution.

\refTbl{Main}, bottom part shows the quantitative results and \refFig{SegmentationResult} the qualitative outcome.
On average, we achieve an DSSIM error of .133 for \textsc{PointLight} and .122 for \textsc{EnvironmentMap}.
The greedy method performs worse ($.161$ and $.154$), as it segments highlights into individual parts.
While the method can understand that highlights belong ``to the rest'' of a material, sometimes they end up in different clusters, as seen in \refFig{FailureModes}, middle and less in the bottom of  \refFig{SegmentationResult}.

\mycfigure{FailureModes}{
Failure modes for all three tasks: blurry highlights, split highlight segmentation and a overshooting DAM.
}

\myfigure{SegmentationResult}{
Results of joint material segmentation and estimation for two samples \textbf{(rows)}.
In every part we show a re-synthesis, as well as two estimated materials and the resulting mask.
The insets in the last row show that, while not all reflection details are reproduced, ours is free of color shifts around the highlights and mostly a low-frequency approximation of the environment reflected.
}

\mysubsection{Discussion}{Discussion}

Typical failure modes are show in \refFig{FailureModes}.
For representation \refFig{FailureModes}, left, the network can overshoot \eg become darker than desired, for unobserved directions.
More input images or a more effective GAN could suppress this.
Sharp details cannot be represented by the cascade of functions of a small network.
Fitting a network with more parameters might be required.
For learning-to-learn \refFig{FailureModes}, right, SGD might produce the right network, but the learned network overshoots.
Similarly, highlights tend to be more blurry (not shown).
For segmentation \refFig{FailureModes}, middle, the rim highlight in the back of the character is purely white and apparently does not look enough like other highlights on blue to be understood.
Consequently, it is assigned the metallic material, which is incorrect.

\begin{wrapfigure}{r}{0.41\linewidth}
\includegraphics[width=\linewidth]{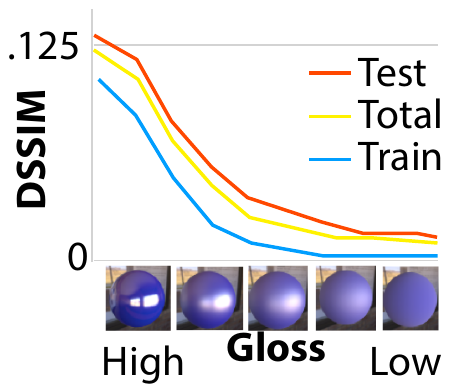}
\caption{Relation of gloss and representation error. 
}
\label{fig:GlossVsError}
\end{wrapfigure}
When the material approaches an arbitrarily complex illumination seen in a mirror, no network can capture all 4D variation anymore.
This relation is shown in the inset plot (\refFig{GlossVsError}) where the vertical axis denotes error, which is decreasing when specular is decreased as well, along the horizontal axis.

\mysubsection{Other Applications}{Applications}
DAMs can also be used for other tasks such as super-resolution, where we extract a DAM in low pixel resolution that can be transferred to a high resolution normal image and denoising of Monte Carlo path tracing, where we extract a DAM from noisy observations and re-generate the image from the DAM, removing the noise.
Detailed evaluation, comparing to a state-of-the-art MC denoiser \cite{bitterli16nonlinearly} and super-resolution \cite{lim2017enhanced} are found in the supplemental material.

\mysection{Real-world Evaluation}{RealWorld}
We have collected a second dataset of photographs of spherical objects with complex appearance (glossy objects under natural light).
In particular we use 3 different materials and 5 different illuminations each with 5 images from registered views (\refFig{RealData}) for training and 70 for testing.
Please see the supplemental video for the animation.
Transfer of appearance captured from a real world image sequence to other complex geometry is shown in \refFig{ShapeTransfer}.

\myfigure{RealData}{
Real-world photo data and our reconstruction (from other views) of multiple materials (denoted \textsf M) in multiple illumination (\textsf L) from multiple views (\textsf V).}

\refTbl{RealWorld} summarizes the outcome for the representation task previously explored for synthetic data only.
An example result is seen in \refFig{Teaser}, more are shown in the supplemental materials.
Our method can estimate view-dependent appearance, unlike RM/RM++, from a small training set, but it can't fully reconstruct mirror-like reflections.

\begin{table}
\center
\caption{
DSSIM (less is better) error on real data.
}
\label{tbl:RealWorld}
\setlength{\tabcolsep}{3.0pt}
\begin{tabular}{l ccccc}
&
\multicolumn{2}{c}{Same view}&
\multicolumn{3}{c}{Novel view}
\\
\cmidrule(lr){2-3}
\cmidrule(lr){4-6}
&\textsc{Our}
&\textsc{RM++}
&\textsc{Our}
&\textsc{RM++}
&\textsc{RM}
\\
\toprule
DSSIM Error
&.069
&.001
&.079
&.090
&.127
\\
\bottomrule
\end{tabular}
\end{table}

\myfigure{ShapeTransfer}{Transfer of appearance from a real video sequence (\textbf{left}) to new 3D shapes (\textbf{right}).}

\mysection{Discussion and Conclusion}{Conclusion}

We have proposed and explored a novel take on appearance processing that neither works on pixel-level IBR-like representations nor by extracting classic explicit reflectance and illumination parameters.
Instead, we work on a deep representation of appearance itself, defined on a generalization of reflectance maps that works in world space where observations cover all directions.
We have shown to enables effective reproduction, estimation by learning-to-learn and joint material estimation-and-segmentation.

In future work, we would like to generalize our approach to  allow independent control of illumination and reflectance (BRDF) \cite{Georgoulis2017Around,Liu2017,Kim2017,deschaintre2018singleimage}, providing an improved editing experience.
Equally, we have not yet explored the symmetric task of learning - how to learn segmenting appearance (\refSec{LearningToLearn}).

\clearpage

{\small
\bibliographystyle{ieee_fullname}
\bibliography{main}
}

\end{document}